%% file: acl_latex.tex
\pdfoutput=1

\documentclass[11pt]{article}

\usepackage[preprint]{acl}

\usepackage{times}
\usepackage{latexsym}

\usepackage[T1]{fontenc}

\usepackage[utf8]{inputenc}

\usepackage{microtype}

\usepackage{inconsolata}

\usepackage{graphicx}

\usepackage{hyperref}

\usepackage{subcaption}

\title{
\raisebox{-0.2\height}{\includegraphics[height=\baselineskip]{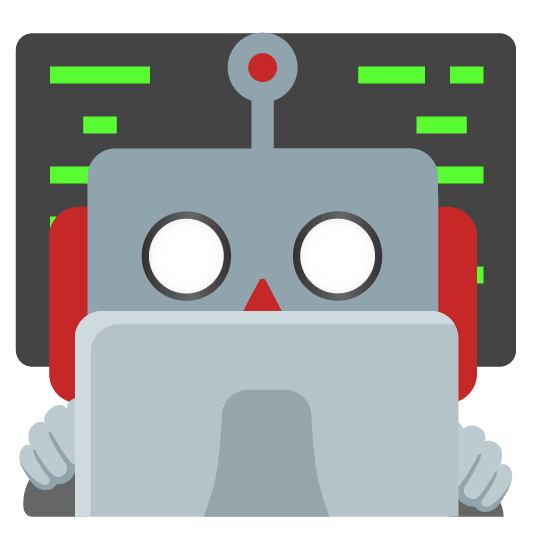}} 
Towards Rationality in Language and Multimodal Agents: A Survey
}

\author{Bowen Jiang\textsuperscript{1, 3}, 
Yangxinyu Xie\textsuperscript{1, 3}, 
Xiaomeng Wang\textsuperscript{1},
Yuan Yuan\textsuperscript{1},
Zhuoqun Hao\textsuperscript{1}, \\ 
{\bf
Xinyi Bai\textsuperscript{2},
Weijie J. Su\textsuperscript{1}, 
Camillo J. Taylor\textsuperscript{1},
Tanwi Mallick\textsuperscript{3}} 
\\  
\begin{tabular}{ccc}
University of Pennsylvania\textsuperscript{1} & Cornell University\textsuperscript{2} & Argonne National Laboratory\textsuperscript{3} \\
Philadelphia, PA, 19104, USA & Ithaca, NY, 14850, USA & Lemont, IL, 60439, USA
\end{tabular}
\\
{\tt\small \{bwjiang@seas, xinyux@wharton, xwang1@wharton, yyuan86@seas, zhuoqunh@sas\}.upenn.edu}
\\
{\tt\small xb52@cornell.edu, \{suw@wharton, cjtaylor@seas\}.upenn.edu, tmallick@anl.gov}
}

\begin{document}
\maketitle
\input{sections/abstract}
\input{sections/introduction}

\input{sections/scope}

\input{sections/preliminary}

\input{sections/main}

\input{sections/evaluations}
\input{sections/conclusion}
\bibliography{custom}

\end{document}

%% file: sections/abstract.tex
\begin{abstract}

This work discusses how to build more rational language and multimodal agents and what criteria define rationality in intelligent systems.
Rationality is the quality of being guided by reason, characterized by decision-making that aligns with evidence and logical principles. It plays a crucial role in reliable problem-solving by ensuring well-grounded and consistent solutions. Despite their progress, large language models (LLMs) often fall short of rationality due to their bounded knowledge space and inconsistent outputs. In response, recent efforts have shifted toward developing multimodal and multi-agent systems, as well as integrating modules like external tools, programming codes, symbolic reasoners, utility function, and conformal risk controls rather than relying solely on a single LLM for decision-making. This paper surveys state-of-the-art advancements in language and multimodal agents, assesses their role in enhancing rationality, and outlines open challenges and future research directions.
We maintain an open repository at \href{https://github.com/bowen-upenn/Agent_Rationality}{https://github.com/bowen-upenn/Agent\_Rationality}.

\end{abstract}

%% file: sections/introduction.tex
\section{Introduction}\label{sec: intro}

\begin{figure*}[t]
  \centering
    \includegraphics[width=\textwidth]{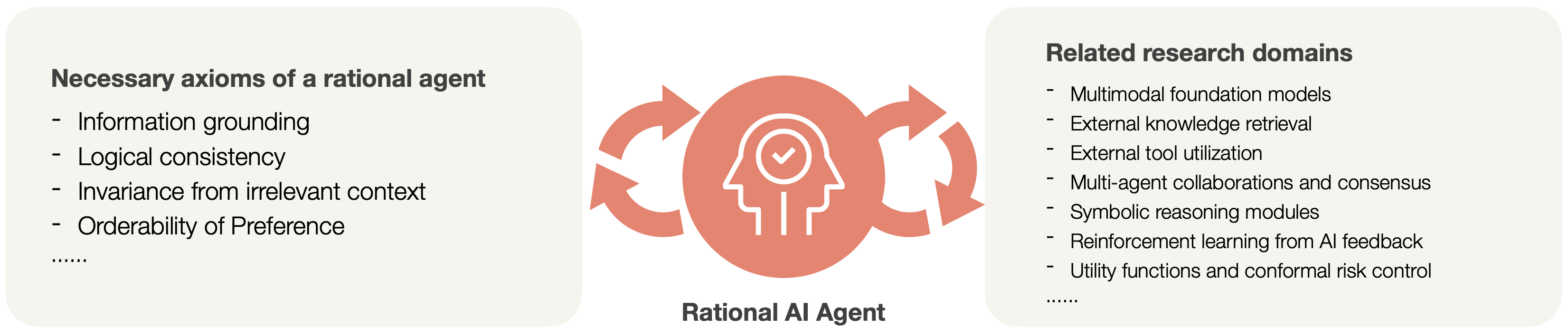}
    \caption{This survey identifies four necessary, though not sufficient, axioms that a rational agent should fulfill. Meanwhile, we reinterpret various research domains related to agents and agent systems through the lens of rationality, examining how their underlying algorithms contribute to each of these axioms.}
    \label{fig:header}
\end{figure*}


Rationality remains a critical and urgent topic in the research of artificial intelligence, particularly as intelligent systems become increasingly involved in high-stakes decision-making processes, such as healthcare, finance, science, and legal services~\cite{he2023survey, li2023large, xie2024wildfiregpt, kang2023deficiency, cheong2024not} where reliability is paramount for human end-users utilizing these agents for decision-making. \textbf{Unlike \textit{reasoning} that aims to draw conclusions from premises, \textit{rationality} ensures that those conclusions are reliably consistent, have an orderability of preference, and are aligned with evidence from various sources and logical principles. }

However, recent studies reveal that even state-of-the-art large language models (LLMs) exhibit limitations in rationality. Because a single LLM relies solely on its internal parametric representations of textual knowledge, while lacking real-world grounding and feedback mechanisms necessary to develop rationality~\citep{bubeck2023sparks, sun2024can, panickssery2024llm}, it shows bounded knowledge, inconsistent responses, and susceptibility to biases and framing effects~\cite{jiang2024peek, chen2024premise, binz2023using, echterhoff2024cognitive, mukherjee2024heuristic, macmillan2024ir, wang2024will, suri2024large}. These limitations raise concerns about their practical reliability in critical sectors, highlighting the need for more reliable and coherent systems capable of rational behaviors. 

To address these challenges, research is shifting toward multimodal agents and multi-agent systems, as complex problems in real life often require collaborations of experts across fields~\cite{eisenfuhr2010rational} and data from diverse sources. 
Formally, we refer \textbf{"agent"}~\cite{bommasani2021opportunities} as an artificial intelligent entity that perceives and understands its environment through various inputs — either natural language or multimodal information like vision, audio, and codes — and acts to achieve specific goals or tasks within natural language domains, while the term "agentic" to describe such behaviors~\cite{kapoor2024ai, Ng2024}. This can encompass a range of systems, from a single LLM, a multimodal foundation model with instruction following capabilities~\cite{liu2024visual, liu2023improved}, to a multi-agent system that integrates multiple AI agents, traditional machine learning or symbolic reasoning modules, external knowledge bases, and tools working together towards a collective goal within the same environment. 

Given this, this survey explores how current literature helps address the limitations of a vanilla LLM in achieving rationality, with an emphasize on language and multimodal agents or agent systems.
We first delineate four necessary, though not sufficient, axioms of rationality Section~\ref{sec:pre} that a rational agent should fulfill, and discuss how current works help move towards each of the axioms in Section~\ref{sec:main}, providing a unique lens to reinterpret their underlying motivations. Lastly, Section~\ref{sec:eval} highlights the lack of sufficient evaluation metrics and benchmarks in the existing literature to adequately measure the rationality of agents, and Section~\ref{sec: open} discusses further open problems. We hope this survey can inspire further research at the intersection between agent systems and cognitive science.

%% file: sections/scope.tex
\section{Scope}\label{sec:scope}

\begin{figure*}[t]
  \centering
    \includegraphics[width=\textwidth]{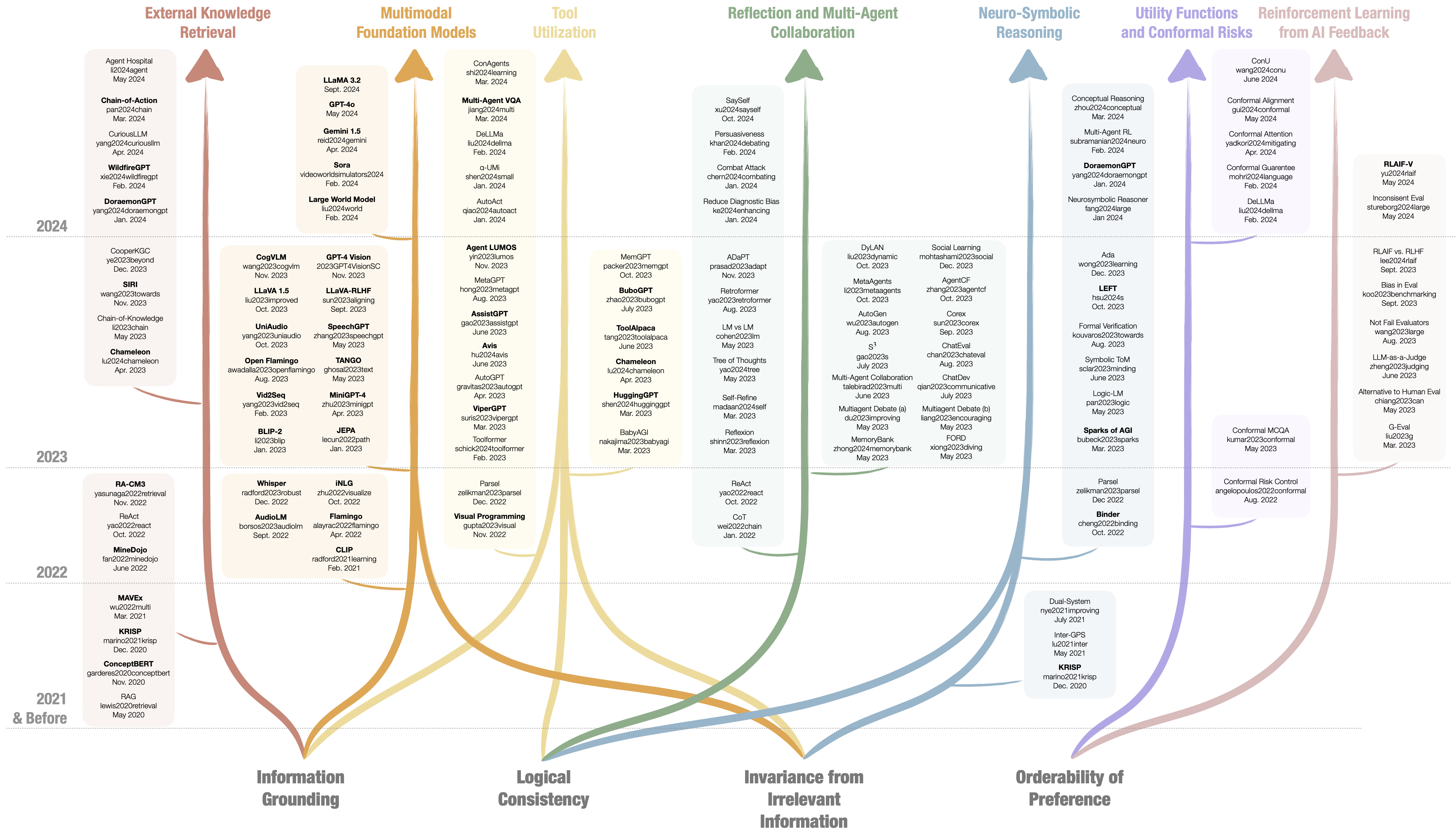}
    \caption{The evolutionary tree of language and multimodal agents and agent systems related to the four key axioms of agent rationality. The axioms are listed at the bottom, while each colored arrow representing a distinct research domain. Works involving multi-modalities are highlighted in \textbf{bold}. }
    \label{fig:tree}
\end{figure*}

Existing surveys in the field of agent systems, such as those on multimodal~\cite{xie2024large, durante2024agent, cui2024survey, xu2024survey, li2024multimodal} and multi-agent systems~\cite{han2024llm, guo2024large, zhang2024llm, cheng2024exploring}, primarily focus on their components, architectures, profiling, planning, communication strategies, memory mechanisms, and applications. Additionally, many works \cite{jiang2024peek, wei2022chain, yao2024tree, cai2024t, valmeekam2023planning, xu2024can, prasad2023adapt, khardon1997learning, huang2022towards, zhang2024llm, qiao2022reasoning} and surveys~\cite{qiao2022reasoning, huang2022towards, ahn2024large, giadikiaroglou2024puzzle, liang2024internal, zheng2024attention, zhang2024llm, xiong2024large} explore the reasoning capabilities of LLMs. Although reasoning plays an important role in ensuring rationality, especially in complex scenarios, it remains parallel to our focus, as mentioned in Section~\ref{sec: intro}. Furthermore, some works touch the aspect of rationality in LLMs~\cite{kassner2023language, raman2024steer, macmillan2024ir}, but they focus on one specific algorithm or application domain. \textbf{To the best of our knowledge, this survey is the first to comprehensively explore the notion of rationality in language and multimodal agents.} We aim to bridge the gap between rationality and agent system, analyzing how designs in these agents and agent systems contribute to advancing certain key axioms of rationality.

%% file: sections/preliminary.tex
\section{Defining Rationality in Agents}\label{sec:pre}
Drawing on foundational works in cognitive science about rational decision-making~\citep{tversky1988rational, hastie2009rational, eisenfuhr2010rational}, this section presents four necessary, though not sufficient, axioms we expect a rational agent or agent systems to fulfill:

\paragraph{Information Grounding}
A rational agent's decision-making should be grounded in physical and factual reality, incorporating information it perceives from multimodal formats and sources. In contrast, an irrational agent generates hallucinations~\cite{huang2023survey}, producing false or misleading information that is not grounded in facts.

\paragraph{Logical Consistency}
Logical consistency refers to an agent's ability to avoid self-contradictions in reasoning and ensure that its conclusions logically follow from its premises. A rational agent should deliver consistent decisions in its final responses, producing invariant decisions across equivalent representations of the same problem.

\paragraph{Invariance from Irrelevant Context}
A rational agent should not be swayed by irrelevant contextual information, focusing instead on the logical essence of the problems and relevant data.

\paragraph{Orderability of Preference}
When comparing alternatives in a decision scenario, a rational agent should be able to rank the options based on the current state and ultimately select the most preferred one based on the expected outcomes.

%% file: sections/main.tex
\section{Towards Rationality in Agents}\label{sec:main}
This section discusses how existing language and multimodal agent systems are advancing the concept of rationality. 
We survey a range of research domains, reinterpret their contributions through the lens of the necessary axioms of rationality outlined earlier, and present a novel perspective that bridges existing methodologies with rational principles.


\subsection{Advancing Information Grounding}
\subsubsection{Grounding on multimodal information}
Grounding an agent solely based on language can be challenging. As a picture is worth a thousand words, recent advances in large multimodal models~\cite{li2024multimodal} integrate language, vision, and other sensory modalities to offer a more comprehensive grounding of information, thereby enhancing the understanding of decision-making contexts. Multimodal foundation models, including but not limited to CLIP~\cite{radford2021learning}, VLBERT and ViLBERT~\cite{su2019vl, lu2019vilbert}, BLIP-2~\cite{li2023blip}, UniAudio~\cite{yang2023uniaudio}, AudioLM~\cite{borsos2023audiolm}, 
TANGO~\cite{ghosal2023text}, SpeechGPT~\cite{zhang2023speechgpt}, (Open) Flamingo~\cite{alayrac2022flamingo, awadalla2023openflamingo}, LLaVA~\cite{liu2024visual, liu2023improved}, CogVLM~\cite{wang2023cogvlm}, MiniGPT-4~\cite{zhu2023minigpt}, Whisper~\cite{radford2023robust}, GPT-4 Vision~\cite{2023GPT4VisionSC} and GPT-4o~\cite{openai2024gpt4o}, LLaMA 3.2~\cite{meta2024llama32}, and Gemini 1.5 Pro~\cite{reid2024gemini} serve as the cornerstones for downstream tasks in multimodal agent systems. More agentic systems increasingly depend on multimodal information to enhance multimodal reasoning.~\cite{zhang2024insightsee, brienza2024multi, elhenawy2024visual, chen2023towards, dong2024insight, wu2024symbol}.

The adaptation of Reinforcement Learning from Human Feedback (RLHF)~\cite{2024o1SC, stiennon2020learning, ouyang2022training, bai2022training, zhang2024prototypical}, a technique popularized in language-only models, also demonstrates promising advancements in reducing hallucination from cross-modal misalignment~\cite{sun2023aligning}. Visual instruction-tuning~\cite{liu2024visual, dai2024instructblip, bai2023qwen, wang2023see} also enables foundation models to engage in more detailed multi-round, context-aware human-agent interactions and collaborations with other agents. This opens the possibility of the System 2 process~\cite{kahneman2011thinking} in multimodal models.

Multi-modalities help expand the functionality of agents by allowing them to access more comprehensive and diverse data.
For example, Chain-of-Action~\cite{pan2024chain} advances the single-modal Search-in-the-Chain~\cite{xu2023search} by supporting multimodal data retrieval for faithful question answering. multimodal understanding in DoraemonGPT~\cite{yang2024doraemongpt} is necessary for spatial-temporal videos analysis. LogicVista~\cite{xiao2024logicvista} expands logical reasoning capabilities to visual contexts. The multimodal capabilities also allow HuggingGPT~\cite{shen2024hugginggpt}, Agent LUMOS~\cite{yin2023lumos}, ToolAlpaca~\cite{tang2023toolalpaca}, and AssistGPT~\cite{gao2023assistgpt} to expand the scope of tasks they can address, including cooperation among specialized agents or tools capable of handling different modalities. Web agents like~\cite{zheng2024gpt, shen2024small, deng2024mind2web, gur2023real, zhou2023webarena, koh2024visualwebarena} grounded on the graphical user interface (GUI) offers higher information density compared to solely HTML codes in textual formats~\cite{yao2022webshop, nakano2021webgpt}.

\subsubsection{Expanding working memory from external knowledge retrieval and tool utilization}

\begin{figure*}[t]
  \centering
  \begin{subfigure}[t]{0.48\linewidth}
    \centering
    \includegraphics[width=\linewidth]{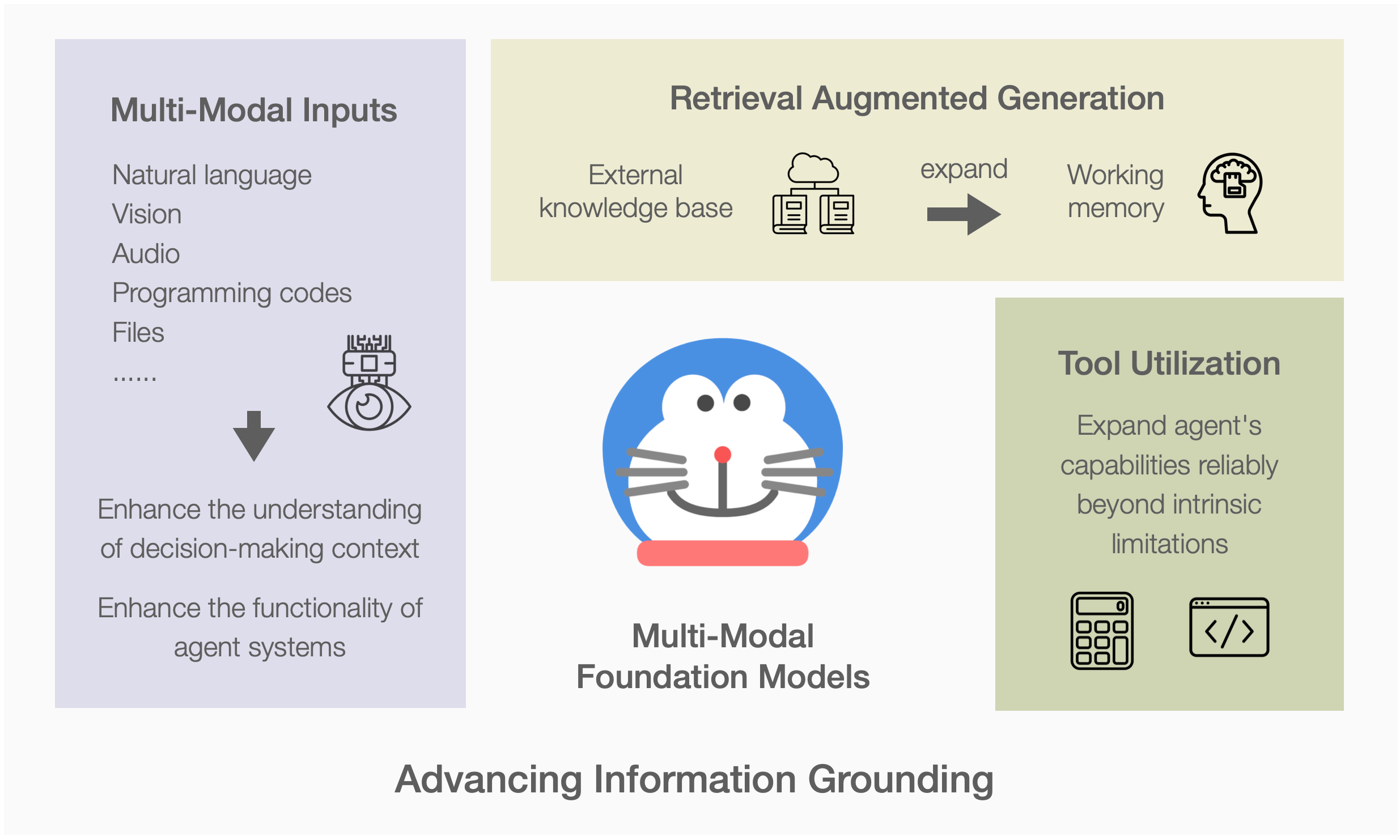}
    \label{fig:info_grounding}
  \end{subfigure}
  \hfill
  \begin{subfigure}[t]{0.48\linewidth}
    \centering
    \includegraphics[width=\linewidth]{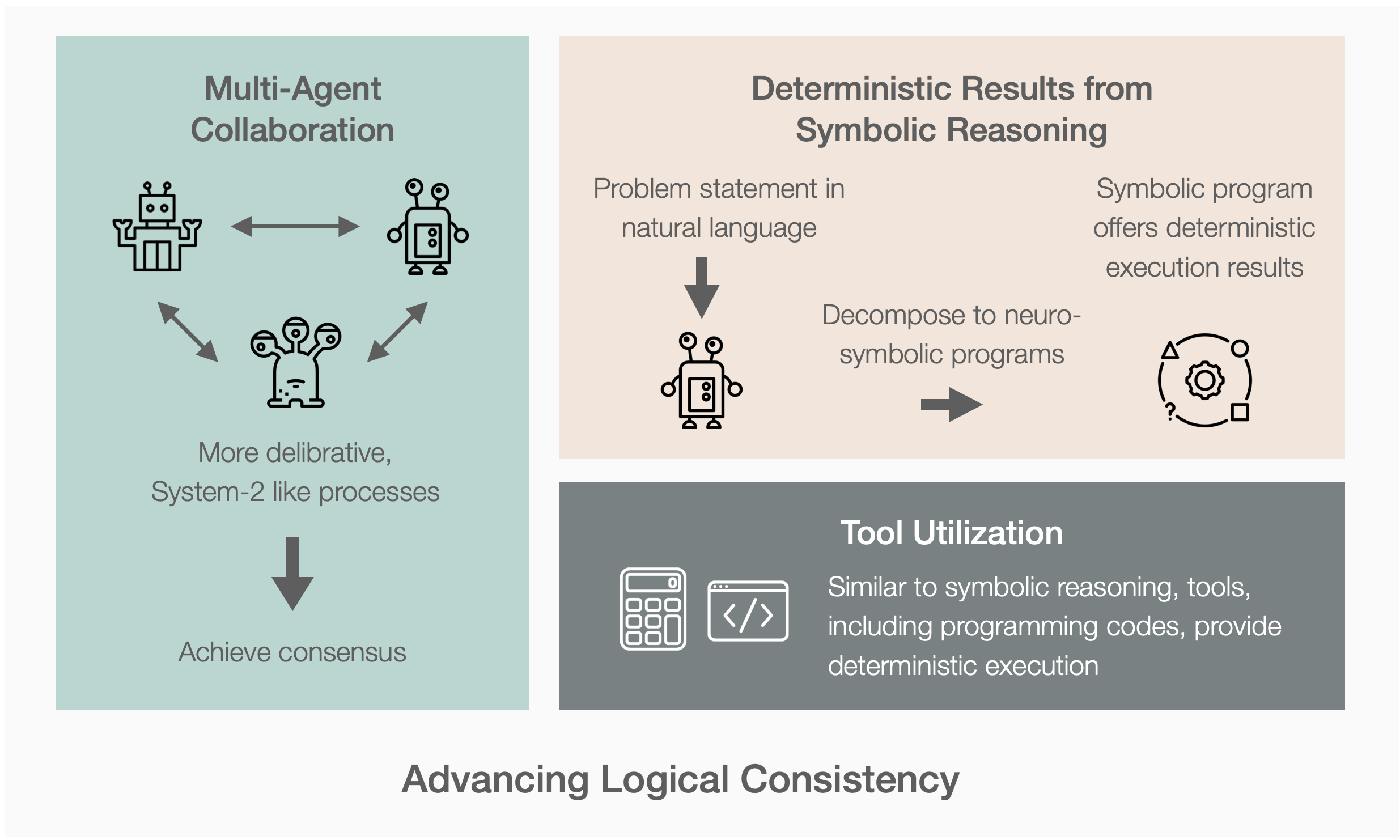}
    \label{fig:logical_consistency}
  \end{subfigure}

  \begin{subfigure}[t]{0.48\linewidth}
    \centering
    \includegraphics[width=\linewidth]{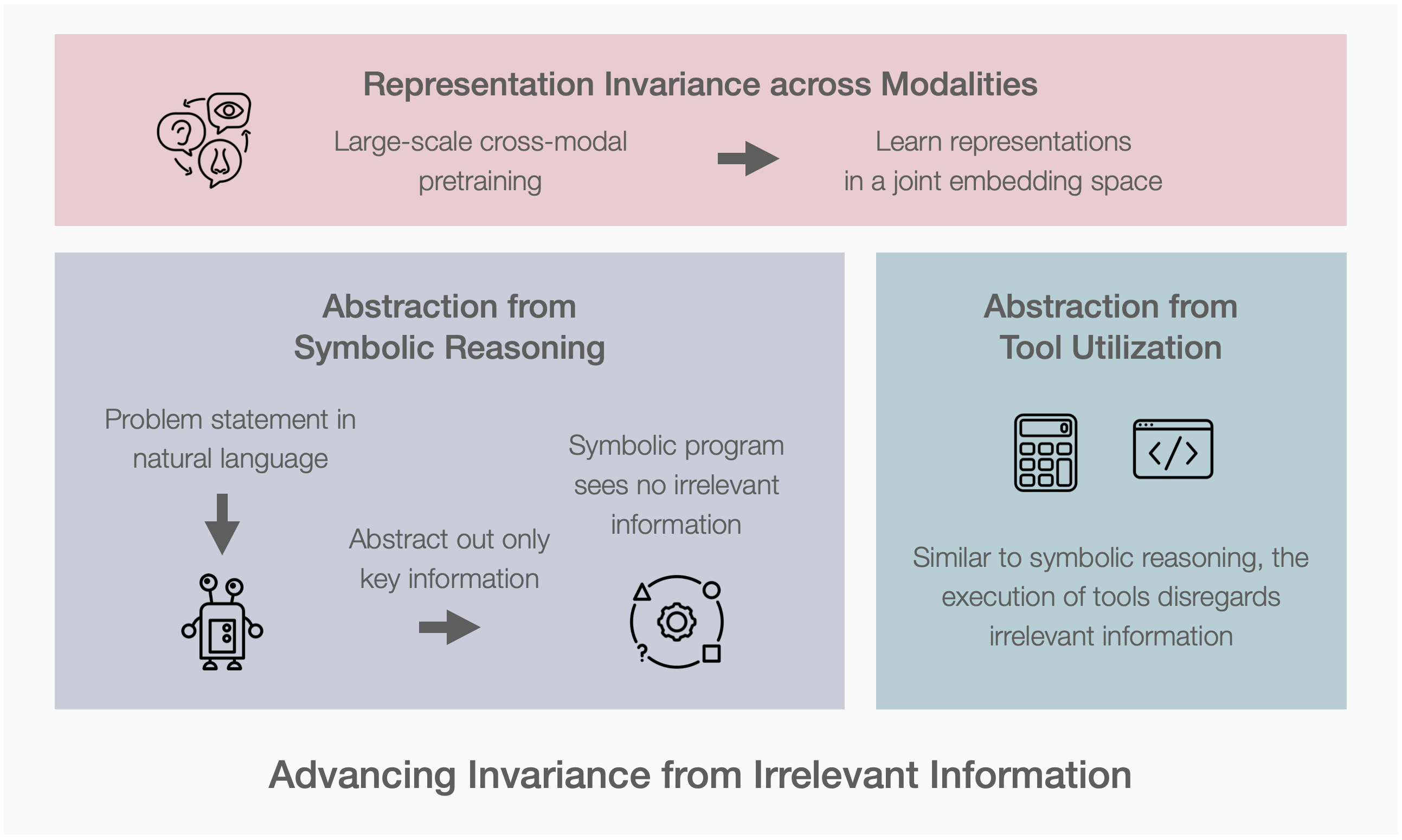}
    \label{fig:invariance}
  \end{subfigure}
  \hfill
  \begin{subfigure}[t]{0.48\linewidth}
    \centering
    \includegraphics[width=\linewidth]{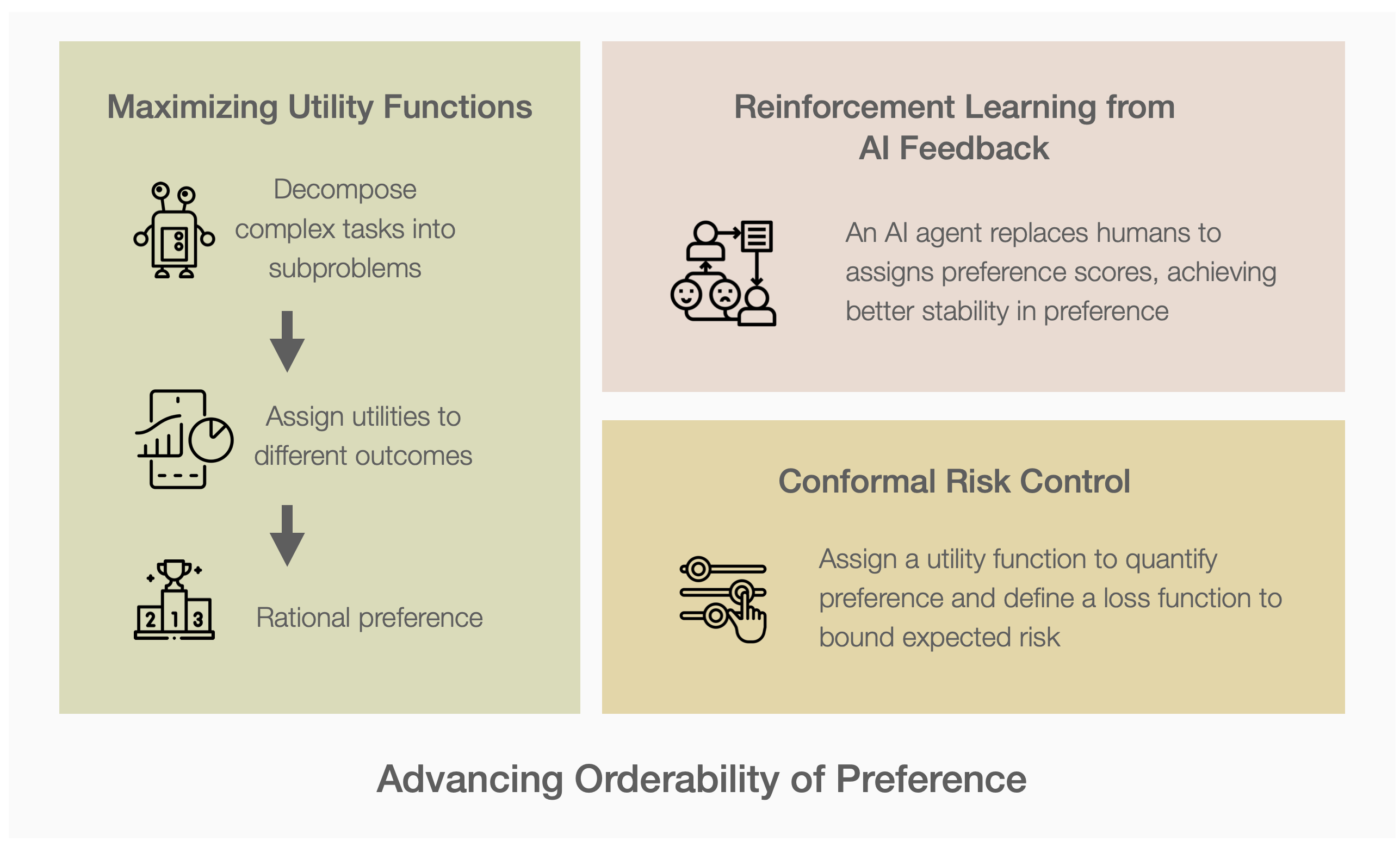}
    \label{fig:goal}
  \end{subfigure}
  \caption{Overview of how language and multimodal agents promote the four axioms of rationality. (1) Top Left - Advancing Information Grounding: Multimodal inputs enhance an agent's understanding of decision contexts and expand its functionalities; External knowledge sources and tools like programming codes expand its bounded working memory. (2) Top Right – Advancing logical consistency: Multi-agent collaboration facilitates deliberate thinking that could correct errors and achieve consensus; Neuro-symbolic reasoning and tools ensure consistent, deterministic executions. (3) Bottom Left – Advancing invariance from irrelevant information: Cross-modal training unifies representations across modalities; Neuro-symbolic tools focus the agent on logical essence. (4) Bottom Right – Advancing orderability of preference: Reinforcement from AI feedback mimics humans and provides more stable preference scores; Utility functions and conformal risk control further guide the preference in rigorous frameworks.}
  \label{fig:four_axioms}
\end{figure*}

Bounded Rationality~\cite{march1958organizations, selten1990bounded} is a concept tailored to cognitively limited agents, suggesting that decision-making is limited by the resources available at hand, and any deviations from the optimal are primarily due to insufficient computational capacity and bounded working memory.

In terms of LLMs, the parametric nature of their existing architecture~\cite{vaswani2017attention} fundamentally limits how much information they can hold. As a result, in the face of uncertainty, LLMs often hallucinate ~\cite{bang2023multitask, guerreiro2023hallucinations, huang2023survey}, generating outputs that are not supported by the factual reality of the environment. 
Retrieval-Augmented Generation (RAG)~\cite{lewis2020retrieval} marks a significant milestone in addressing such an inherent limitation of LLMs. 
Broadly speaking, RAG refers to any mechanism that provides external knowledge to the input context of an LLM and helps it deliver responses with up-to-date, factual, and grounded information, especially in scientific and medical domains. Examples include Chameleon~\cite{lu2024chameleon}, WildfireGPT~\cite{xie2024wildfiregpt}, and Agent Hospital~\cite{li2024agent}, TILP~\cite{xiong2023tilp}, as well as Chain-of-Knowledge~\cite{li2023chain} which finds that integrating multiple knowledge sources enhances performance by $2.1\%$ compared to using a single source.
Another line of systems construct large-scale knowledge graphs~\cite{hogan2021knowledge, he2024givestructuredreasoningknowledge} from real-world sources to effectively expand their working memory. 

Enabling agents to use tools also expands their bounded working memories and grounds their responses by the outputs of these tools. Toolformer~\cite{schick2024toolformer} opens a new era that allows LLMs to use external tools, effectively extending their capabilities beyond intrinsic limitations.
A multi-agent system can coordinate agents understanding when and which tool to use, which modality of information the tool should expect, how to call the corresponding API, and how to incorporate outputs from the API calls, which anchors subsequent processes with more accurate information beyond their parametric memory. For example, VisProg~\cite{gupta2023visual}, ViperGPT~\cite{suris2023vipergpt}, and Parsel~\cite{zelikman2023parsel} generate Python programs to reliably execute subroutines. \citet{xiong2024search} extends LLMs with search engines. \citet{gupta2023visual, suris2023vipergpt} also invoke off-the-shelf models for multimodal assistance. These systems no longer need to generate all responses from scratch, instead relying on tools for more accurate and reliable information.

\subsection{Advancing Logical Consistency}
\subsubsection{Consensus from reflection and multi-agent collaboration}
"Thinking, fast and slow"~\cite{kahneman2011thinking} defines System 1 and System 2 as two types of thinking processes in human cognitive systems. System 1 is fast, intuitive, and automatic, often relying on heuristics, while System 2 is slower, more deliberate, and analytical, engaging in logical reasoning. Due to the probabilistic outputs of LLMs, which resemble the fast, non-iterative nature of System 1 thinking. In contrast, multi-agent systems that promote debate and consensus among AI agents can help align outputs more closely with the slow, deliberate decision-making typical of System 2 processes, thus enhancing logical consistency. 

Multi-round self-reflection that encourages a single LLM to critically evaluate its previous responses has demonstrated the effectiveness of deliberation in improving logical coherence  and qualities~\cite{2024o1SC, shinn2024reflexion, madaan2024self, wang2022self, zhong2024memorybank, lu2023memochat, xu2024sayself, xu2024can, xiong2024deliberate}. Expanding on this, multi-agent systems introduce collaboration among multiple agents~\cite{zhang2024survey}, enabling collective deliberation through cross-examination and debate. For instance, LM vs LM~\cite{cohen2023lm} introduces a cross-examination between two agents to detect errors and make factuality decisions. FORD~\cite{xiong2023diving} reports an accuracy increase up to $4.9\%$ compared to a single LLM. AgentReview~\cite{jin2024agentreview} presents how discussions cause distribution shifts in final reviews compared to initial reviews. \citet{liang2023encouraging} demonstrates the superiority of multi-agent debate over self-reflection, with final consensus achieving a $16.0\%$ improvement in reasoning tasks. \cite{zhao2023competeai} investigates the multi-agent competition behaviors. Similarly, \citet{du2023improving} finds that LLMs can converge on a single shared answer after multiple rounds of debate, resulting in a factual accuracy increase of $7.2$-$15.9\%$ across tasks.
All these approaches enhance the system's capability to capture initial errors, improve factuality in reasoning, and achieve final consensus with fewer inconsistencies.

\subsubsection{Consistent execution from symbolic reasoning and tool utilization}
Neuro-symbolic reasoning~\cite{zelikman2023parsel, pan2023logic, sclar2023minding, hsu2024s, fang2024large, yang2024doraemongpt, subramanian2024neuro} combines learning abilities with symbolic systems for explicit knowledge representation and logical reasoning. A multi-agent system incorporating symbolic modules can not only understand language queries but also solve them with a level of consistency, providing a faithful and transparent reasoning process based on well-defined rules that adhere to logical principles, which is unachievable by a single LLM. For example, LEFT~\cite{hsu2024s} uses "left" as an quintessential example of concepts in multimodal models. It demonstrates how multimodal models can generate first-order logic programs to reason about domain-specific relational concepts. Logic-LM~\cite{pan2023logic}, KRISP~\cite{marino2021krisp}, Binder~\cite{cheng2022binding}, Parsel~\cite{zelikman2023parsel}, and \citet{fang2024large} also utilize symbolic modules or graphs to deliver consistent outputs. 

Similarly, tool utilization abstracts problems into deterministic tool executions, such as calculators, calendars~\cite{schick2024toolformer}, and programming codes, and seamlessly integrate reliable outputs back to the responses. For instance, ToolAlpaca~\cite{tang2023toolalpaca} generates a tool-use corpus in a multi-agent simulation environment. Binder~\cite{cheng2022binding} converts queries into Python or SQL codes to interact deterministically with structured knowledge bases. Parsel~\cite{zelikman2023parsel} combines a code-LLM with a constraint solver to deterministically handle decomposed tasks. VisProg~\cite{gupta2023visual}, ViperGPT~\cite{suris2023vipergpt}, and Parsel~\cite{zelikman2023parsel} generate Python programs to execute subroutines. Besides, BabyAGI~\cite{nakajima2023babyagi}, Chamelon~\cite{lu2024chameleon}, AssistGPT~\cite{gao2023assistgpt}, Avis~\cite{hu2024avis}, ToolAlpaca~\cite{tang2023toolalpaca}, MetaGPT~\cite{hong2023metagpt}, Agent LUMOS~\cite{yin2023lumos}, AutoAct~\cite{qiao2024autoact}, $\alpha$-UMi~\cite{shen2024small}, and ConAgents~\cite{shi2024learning} harness compositional reasoning to enable modular tool-using capabilities in real-world scenarios.
All these examples demonstrate how deterministic execution, whether through symbolic reasoning or tool utilization, ensures consistent outcomes, making complex reasoning processes more robust and transparent.

\subsection{Advancing Invariance from Irrelevant Information}
\subsubsection{Representation invariance across modalities}
Given adequate information grounding, agents should make consistent decisions across different modalities that share equivalent underlying logic. Multimodal foundation models are particularly adept at promoting invariance by processing multimodal data in an unified representation. Specifically, their large-scale cross-modal pretraining stage seamlessly tokenizes both vision and language inputs into a joint hidden embedding space, learning cross-modal correlations through a data-driven approach. In other words, image tokens are simply regarded as a foreign language~\cite{wang2022image}. Moreover, the cross-modal validation inherent in multimodal foundation models allows for reconciliation of data from different modalities, closing their distance in the hidden embedding space~\cite{radford2021learning}.

The concept of invariance is the cornerstone of Visual Question Answering (VQA) agents~\cite{chen2022pali, jiang2024multi, wang2023towards, yi2018neural, wang2022image, bao2022vlmo, li-24-vqa, zhao2023less, fan2024muffin}. On one hand, these agents must grasp the invariant semantics of any open-ended questions posed about images, maintaining consistency despite variations in wording, syntax, or language. On the other hand, within a multi-agent VQA system, visual agents can provide crucial verification and support for language-based reasoning~\cite{wang2023towards, jiang2024multi, zhao2023less}, while language queries can direct the attention of visual agents, based on a shared and invariant underlying knowledge across vision and language domains.

\subsubsection{Abstraction from symbolic reasoning and tool utilization}
In most cases, tools or symbolic reasoners require translating natural language queries into function calls with predefined syntax. Once the function calls and their input arguments are determined, the tools or symbolic reasoners will narrow down their focus to logical essense, ignoring any irrelevant context in the original queries as long as they are logically equivalent. For example, in Multi-Agent VQA~\cite{jiang2024multi}, a language model extracts only the relevant object names and passes them to Grounded SAM~\cite{ren2024grounded}, an object detection tool, instead of sending the entire visual question. Similarly, LEFT~\cite{hsu2024s} abstracts target objects from complex 3D visual scenes into symbolic representations to predict their relational properties, ensuring that symbolic reasoning is unaffected by other contextual details in the environment. This abstraction enables more focused and efficient reasoning processes.

\subsection{Advancing Orderability of Preference}
\subsubsection{Learning preference from reinforcement learning}
Reinforcement Learning from Human Feedback (RLHF)~\cite{2024o1SC, stiennon2020learning, ouyang2022training, bai2022training} helps reduce the preference gap between agents and humans. However, we argue that RLHF does not guarantee rational preference orderability, as human preferences are often inconsistent and vary across individuals. An emerging alternative is Reinforcement Learning from AI Feedback (RLAIF)~\cite{lee2024rlaif, wang2023large, zheng2023judging, chiang2023can, liu2023g, koo2023benchmarking, stureborg2024large, yu2024rlaif}. RLAIF leverages LLMs as evaluators, achieving more stable preference over different formatting of the task instructions and the sampling algorithm used to generate the answers~\cite{chiang2023can}. Additionally, \citet{yu2024rlaif} demonstrates how a multimodal agent can iteratively assign trustworthiness scores to each atomic claims, further enhancing the reliability of evaluation process.

\subsubsection{Maximizing utility functions and controlling conformal risks}
Recent work also explores the expected utility theory (EUT) \cite{von2007theory} to improve the decision-making capabilities of language models. EUT provides a formal framework to quantify preferences by assigning utility values to outcomes and calculating the expected utility based on the weighting probability of each outcome’s occurrence. For example, DeLLMa~\cite{liu2024dellma} applies this approach by decomposing complex decision problems into subtasks, assigns utilities to different outcomes, and selects actions that maximize expected utility. 

\citet{angelopoulos2022conformal} introduces a conformal risk control framework that ensures the expected loss, under any non-increasing loss function, remains bounded by a predefined threshold $\alpha$. 
This framework has been adapted to control a wide range of metrics, including factuality, false discovery rate, and hallucination frequency \cite{mohri2024language, cherian2024large, kumar2023conformal, wang2024conu, yadkori2024mitigating, overman2024aligning, gui2024conformal}.

%% file: sections/evaluations.tex
\section{Evaluating Rationality in Agents}\label{sec:eval}
While there are numerous reasoning benchmarks \citep{talmor2019commonsenseqa, liu2021logiqa, liu2023logiqa, yang2018hotpotqa, hendrycks2021measuring, chen2020hybridqa, zhou2024conceptual, rasheed2024large, ma2024agentboard, wang2024benchmark, abdelnabi2023llm}, they do not directly measure rationality. The amount of studies for evaluating rationality in agent or agent systems remains scant, despite the growing interest in the field. In this section, we explore potential evaluation methods and benchmarks aligned with each of the proposed axioms of rationality.


\subsection{Evaluating Information Grounding}
Information grounding is usually evaluated by the level of hallucination~\cite{bang2023multitask, guerreiro2023hallucinations, huang2023survey}. 
Multiple evaluation benchmarks targeting language-only dialogue have been proposed, such as BEGIN~\cite{dziri2022evaluating}, HaluEval~\cite{li2023halueval}, DialFact~\cite{gupta2021dialfact}, FaithDial~\cite{dziri2022faithdial}, EureQA~\cite{li2023deceiving}, AIS~\cite{rashkin2023measuring}, and others~\cite{zheng2023does, das2023diving, cao2021hallucinated, liu2024exploring}. However, benchmarks for multimodal agents beyond language dialogue remain limited. Some efforts include POPE~\cite{li2023evaluating}, LLaVA-RLHF~\cite{sun2023aligning}, BLINK~\cite{fu2024blink}, and \citet{rohrbach2018object, biten2022let} are the few examples that consider multimodal hallucination.
The community needs more hallucination benchmarks to quantitatively evaluate the extent to which multi-modality and multi-agents reduce hallucinations in comparison with single LLMs.

\subsection{Evaluating Logical Consistency}
To assess whether LLMs can generate logically consistent responses across different but inherently equivalent framing of the same tasks, studies introduce perturbations to the original task descriptions. Perturbation techniques include modifying instruction templates \cite{weber2023mind}, paraphrasing task descriptions \citep{yang2023rethinking, ohmer2024form}, translating the prompts into a different language \citep{ohmer2023separating, ohmer2024form, xu2024exploring} and back~\cite{yang2023rethinking}, and altering the order of in-context learning exemplars \cite{lu2021fantastically, pecher2024sensitivity}. \citet{jiang2024peek} further highlights inconsistent behavior across state-of-the-art LLMs when faced with token biases, even when the logical essence of the tasks remains intact. 

Furthermore, uncertainty quantification~\cite{lin2023generating, xiao2022uncertainty, ye2024benchmarking, shen2024thermometer, shen2024uncertainty, xiong2023can} provides insights when an agent may produce inconsistent responses, helping improve their robustness.

\subsection{Evaluating Invariance from Irrelevant
Information}
Studies such as \citet{shi2023large}, \citet{wu2024easily}, \citet{liu2024lost}, and \citet{yoran2023making} investigate the phenomenon of ``lost-in-context" by introducing random or misleading sentences into original problem statements. Early benchmarks such as those by \citet{weston2015towards}, \citet{sinha2019clutrr}, \citet{clark2020transformers}, and \citet{webson2021prompt} also incorporate irrelevant content. More recent benchmarks like MileBench \cite{song2024milebench}, Mementos \cite{wang2024mementos}, Seed-bench-2 \cite{li2023seed}, and DEMON \cite{li2023fine} extend these evaluations to multimodal agents acting in long context with image sequences. In these scenarios, the agents must accurately isolate relevant information from large context windows.

\subsection{Evaluating Orderability of Preference}

Having an orderability of preference is essential when leveraging LLMs as evaluators to ensure reliable assessments. 
\citet{luo2023chatgpt, shen2023large, gao2023human, wang2023chatgpt, chen2023exploring, chiang2023closer, zheng2024judging, fu2023gptscore, liu2023gpteval} highlight challenges with LLMs in this role, reporting inconsistent ratings and difficulties in establishing reliable comparisons. This inconsistencies raise concerns about the ability of LLMs to accurately rank and evaluate different options or responses. ChatEval~\cite{chan2023chateval} and \citet{bai2024benchmarking} suggest improved preference aligned with humans through multi-agent collaborations. 
The multiple choice problems~\cite{paperswithcode-mcqa} serves as another common testing ground. \citep{zong2023fool, zheng2023large} show that LLMs are susceptible to the rearranging of options, often failing to maintain a coherent order of preference. 


The expected utility theory \cite{von2007theory} provides a prototypical framework to evaluate an LLM agent's preferences, informed by specific parameters of the utility function. Building on this framework, \citet{jia2024decision} reveals that LLMs exhibit human-like patterns such as risk aversion, loss aversion, and overweighting small probabilities under uncertainty. \citet{ross2024llm} identifies biases like time discounting, reflecting the preference for discounting non-immediate gains. It finds that these agents are neither entirely human-like nor economicus, i.e., rational economic beings, highlighting the need for intervening in their behavior towards better alignment with desired objectives.

%% file: sections/conclusion.tex

\section{Open Problems and Future Directions} \label{sec: open}
\paragraph{Towards Inherent Rationality} 
Despite significant research efforts in this survey aimed at achieving one or more axioms of rationality, most existing algorithms rely on external tools, thereby not inherently enhancing the rationality of artificial intelligence. In other words, current methods are neither sufficient nor necessary to achieve human-level rationality, but they serve as instrumental tools that bridge the gap between an LLM’s response and rationality. These approaches enable agent systems to more closely mimic rational thinking in their output responses from the end-user's perspective. However, how to effectively close the loop and bake these more rational outputs back into foundation models themselves~\cite{zhao2024we} beyond mere fine-tuning remains an open question. It remains a question if we can leverage these more rational outputs, or training the model to verify against rational axioms for reward scores, to \textit{inherently} enhance a single foundation model's rationality in its initial responses without external assistance.

\paragraph{Encouraging More Multimodality in Multi-Agent Systems} Research into the integration of multi-modality within multi-agent systems would be promising. Fields such as multi-agent collaboration and symbolic reasoning, as shown in Figure~\ref{fig:tree}, currently under-utilize the potential of multimodal sensory inputs. We believe that expanding the role of multi-modalities, including but not limited to vision, audio, and structured data could continue enhancing the rationality of multi-agent systems.

\paragraph{Needing More Comprehensive Evaluation of Rationality} 
The choices of evaluation metrics are important~\cite{schaeffer2024emergent}. 
Although there have been some efforts to assess rationality in agent systems, the field still lacks comprehensive metrics. Existing evaluations predominantly focus on the final performance, neglecting the most interesting intermediate steps and different axioms of rationality, and provides limited investigations into multimodal and multi-agent systems. 
A promising direction is to create methods specifically tailored to assess rationality, going beyond existing ones on accuracy. These future methods should account for nuanced token biases~\cite{jiang2024peek}, tolerate perturbations, and avoid data contamination to yield robust, statistically significant results.

\section{Discussion}
Rationality adds another crucial dimension to the performance of intelligent agents besides reasoning capabilities. More rational agents - capable of providing coherent orderability of preferences and grounded in knowledge beyond their bounded parametric memory - could enhance their roles in automatic evaluation and preference alignment, where human involvements are expensive yet unstable. 

It becomes increasingly important for human users applying these agents in critical sectors like health care and finance that expect consistent and reliable decision-making.
For example, in financial domains, decision-making by LLMs must operate within an acceptable risk threshold. By adapting conformal risks, we can enhance the orderability of preference and create a theoretical framework for parameterizing risk preferences into utility models. This approach enables rational financial agents to optimize decisions to max utilities while controlling risks through loss functions that bound expected risks.

Medical domains also present unique multi-modal challenges, particularly in medical imaging interpretation. For instance, diagnostic methods must withstand irrelevant information like background noise in MRI scans. An VLM could provide diagnostic explanations by looking at these images, and an LLM could collaborate with it as a multi-agent system to verify against pre-defined illness criteria, incorporating external knowledge retrieval to improve information grounding. Besides, a multi-agent collaboration could involve diverse medical-domain personas, such as surgeons, nurses, physicians, radiologists, pharmacists, offering diagnostic recommendations from different perspectives, potentially reducing hallucinations through collaborations.
As a result, we hope our survey could serve as a tool-box for agent developers who want to build more rational agents in diverse domains.

This survey investigate current approaches in a range of related literature that advance towards more rational language and multimodal agents. These approaches include the integration of multimodal inputs, reflections and multi-agent collaborations, RLAIF, utility functions, conformal risk controls, and modules that perform deterministic execution like tools - including programming codes - and neuro-symbolic reasoning modules.

We believe in the value of further investigations into the rationality of AI agents, particularly through a collaboration between the AI research community and cognitive psychologists for a deeper understanding of rationality.

\section{Limitations}
The fields of language and multimodal agents are rapidly evolving. Despite our best efforts, it is inherently impossible to encompass all related works within the scope of this survey. Our discussion also possesses limited mention of the reasoning capabilities, theory of mind in machine psychology, cognitive architectures, and rationality models like formal logic and probability theory, all of which lie beyond the scope of this survey but are crucial for a deeper understanding of the agent systems. Furthermore, we present necessary but not sufficient axioms of rationality; no methodologies mentioned in our survey could sufficiently guarantee a genuine rationality in agents. The concept of rationality in human cognitive science may encompass more principles and axioms than those defined in our survey, such as completeness, transitivity, monotonicity, decompoability~\cite{poole2010artificial}, which are more theoretical and fundamental in nature and not directly related to language and multimodal agents discussed in our survey.

\section{Acknowledgement}
This work was funded by the Laboratory Directed Research and Development (LDRD) program at Argonne National Laboratory, with support from the Office of Science, U.S. Department of Energy, under Contract No. DEAC02-06CH11357. This work was also supported by NSF grant CCF-2112665 (TILOS), which provides funding for B. Jiang. and C. J. Taylor. Y. Xie and W. J. Su also acknowledge support from the NSF HDR TRIPODS award (CCF-1934876). 